\documentclass[sigconf]{acmart}
\usepackage{booktabs}
\usepackage{multirow}
\usepackage{graphicx}

\usepackage{amssymb}
\usepackage{textcomp}

\usepackage{subcaption}
\usepackage{array}
\usepackage{float}

\usepackage{tabularx} 
\usepackage{ragged2e} 
\usepackage{enumitem}    % Compact itemize environments
\usepackage{amsmath}   

\usepackage{adjustbox}

\AtBeginDocument{%
  }

\setcopyright{none}
\renewcommand\footnotetextcopyrightpermission[1]{}%
\begin{document}

\title{TeleCom-Bench: How Far Are Large Language Models from Industrial Telecommunication Applications?}

% Can large language models include telecom knowledge or apply in industrial scenarios well?
% how far are current large language models away from application in industrical telecom scenarios?
% Evaluating the telecom knowledge and applicatical ability for large language models

% \author{Anonymous Authors}
% \author{
%   Jieting Xiao\textsuperscript{1},
%   Yun Lin\textsuperscript{1},
%   Huizhen Qiu\textsuperscript{1},
%   Rui Ma\textsuperscript{1},
%   Chen Zhong\textsuperscript{1},
%   Dongyang Xu\textsuperscript{1},
%   Xiao Long\textsuperscript{1},
%   Chaoyu Zhang\textsuperscript{1},
%   Qiaobo Hao\textsuperscript{1},
%   Ding Zou\textsuperscript{1, \dagger, *},
%   Zhiguo Yang\textsuperscript{1},
%   Yanqin Gao\textsuperscript{1},
%   Fang Tan\textsuperscript{1}
% }

% % 机构列表：一个机构一行
% \affiliation{
%   \textsuperscript{1}
%   \institution{ZTE Corporation}
%   \country{China}
% }

% % 邮箱列表：合并同机构邮箱，用 \textsuperscript{编号} 标注归属
% \affiliation{
%   \textsuperscript{1} \{xiao.jieting, lin.yun, qiu.huizhen, ma.rui13, zhong.chen1, xu.dongyang2, long.xiao, zhang.chaoyu, hao.qiaobo, zou.ding, yang.zhiguo, gao.yanqin, tan.fang\}@zte.com.cn
% }

\author{
  Jieting Xiao\textsuperscript{1},
  Yun Lin\textsuperscript{1},
  Huizhen Qiu\textsuperscript{1},
  Rui Ma\textsuperscript{1},
  Chen Zhong\textsuperscript{1},
  Dongyang Xu\textsuperscript{1},
  Xiao Long\textsuperscript{1},
  Chaoyu Zhang\textsuperscript{1},
  Qiaobo Hao\textsuperscript{1},
  Ding Zou\textsuperscript{1, \textdaggerdbl, *},
  Zhiguo Yang\textsuperscript{1},
  Yanqin Gao\textsuperscript{1},
  Fang Tan\textsuperscript{1}
}

% % 机构列表：一个机构一行
\affiliation{
  \textsuperscript{1}
  \institution{ZTE Corporation}
  \country{China}
}

% % 邮箱列表：合并同机构邮箱，用 \textsuperscript{编号} 标注归属
 \affiliation{
  \textsuperscript{1} \{xiao.jieting, lin.yun, qiu.huizhen, ma.rui13, zhong.chen1, xu.dongyang2, long.xiao, zhang.chaoyu, hao.qiaobo, zou.ding, yang.zhiguo, gao.yanqin, tan.fang\}@zte.com.cn
  \country{China}
}

% \acmConference[TeleCom-Bench 2025]{The First Workshop on Telecom Foundation Models}{TBD}{TBD}
\renewcommand{\shortauthors}{Jieting Xiao et al.}
\acmConference[KDD'2026]{}{August 9-13}{2026, Jeju, Korea}

\begin{abstract}
While Large Language Models have achieved remarkable integration in various vertical scenarios, their deployment in the telecommunications domain remains exploratory due to the lack of a standardized evaluation framework. Current telecom benchmarks primarily focus on static, foundational knowledge and isolated "atomic" skills, neglecting the \textbf{\textit{equipment-specific documentation}} and \textbf{\textit{end-to-end industrial workflows}} essential for real-world production systems. 
To bridge this gap, we present TeleCom-Bench, a comprehensive benchmark comprising 12 evaluation sets with 22,678 curated samples, which evaluates LLMs across a synergistic hierarchy: (1) Multi-dimensional Knowledge Comprehension, which integrates telecommunication fundamentals, 3GPP protocols, and 5G network architecture with proprietary product knowledge across wired, core, and wireless networks via knowledge graph-driven synthesis; and (2) End-to-End Knowledge Application, which formalizes six core tasks on authentic trajectories from live network agent workflows, including intent recognition, entity extraction, event verification, tool invocation, root cause analysis, and solution generation—across network optimization and fault maintenance scenarios.
Evaluations of eight state-of-the-art LLMs reveal a universal Execution Wall: while models achieve \textgreater{}90\% accuracy in linguistic interface tasks such as intent recognition and entity extraction, performance collapses to approximately 30\% in procedural execution tasks like solution generation. This capability gap demonstrates that current LLMs function competently as \textit{diagnosticians} but fail as \textit{field engineers}. TeleCom-Bench provides standardized diagnostics to precisely pinpoint this deficit, offering actionable guidance for domain-specific alignment toward production-ready telecom agents. The dataset and evaluation code will be released at \url{https://github.com/ZTE-AICloud/TeleCom-Bench}.

% To bridge this gap, we introduce TeleCom-Bench, the first systematic benchmark designed to evaluate LLMs across a hierarchy of multi-dimensional knowledge comprehension and end-to-end knowledge application, comprising 17 evaluation sets with 22,409 curated samples.
% TeleCom-Bench uniquely integrates proprietary equipment manuals and technical patents with 3GPP standards, employing a dual-synthesis strategy of knowledge graph-driven modeling and distillation-driven extraction. Furthermore, it formalizes authentic engineering trajectories—such as network optimization and fault root-cause analysis—into rigorous, multi-step task chains to assess a model’s operational readiness as a decision-making agent. We conduct an extensive comparative analysis of leading open-source and proprietary LLMs, diagnosing critical failure modes and performance bottlenecks. Our findings provide actionable insights and a rigorous roadmap for transitioning telecom-focused LLMs from academic prototypes to production-ready intelligent systems.
\end{abstract}




\maketitle

\footnotetext{\textdaggerdbl\ Corresponding author.}
\footnotetext{*\ Project Leader.}

\section{Introduction}

The rapid evolution of Large Language Models~\cite{hu2025survey,zhou2025large} has catalyzed a paradigm shift across diverse vertical sectors~\cite{radanliev2025artificial,zhu2025we,shi2025continual}. By aligning general-purpose reasoning with specialized knowledge structures, domain-specific LLMs~\cite{naveed2025comprehensive,song2025injecting,boateng2025survey} have achieved profound integration in fields such as healthcare, finance, and software engineering. These models serve not only as repositories of static knowledge but as core engines driving real-world business logic and automated workflows. However, within the telecommunications sector—a domain characterized by stringent technical standards, heterogeneous operational environments, and highly concentrated expertise—the deployment of LLMs remains in its nascent, exploratory stage~\cite{zhou2024large,jiang2025comprehensive}. A critical bottleneck is the absence of a standardized and comprehensive evaluation framework capable of quantifying whether these models possess the requisite telecom-specific intelligence to function effectively within autonomous agents or workflows~\cite{long2025survey,zou2025telecomgpt}.

\begin{figure}[t]
\centering
\includegraphics[width=0.95\linewidth]{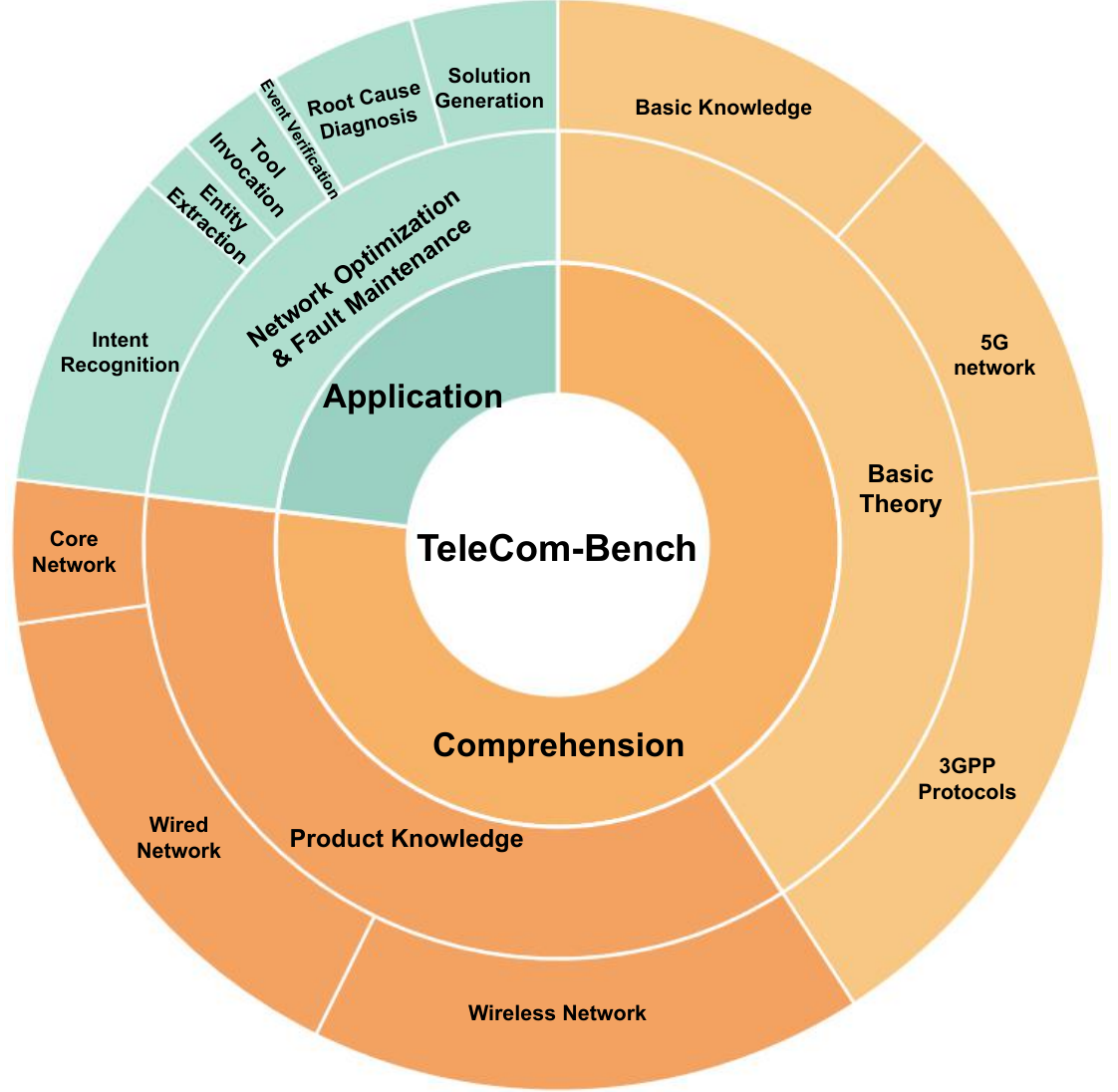}
\caption{Hierarchical Evaluation Structure of TeleCom-Bench: Knowledge Comprehension and Application}
\label{fig:dataset_structure}
\end{figure}

The construction of an effective telecom benchmark necessitates addressing two pivotal dimensions: comprehensive knowledge coverage and scenario-accurate execution. Existing benchmarks, such as SPEC5G and TeleQnA, primarily focus on foundational communication standards and theoretical principles~\cite{zhou2024large}. Others, like TeleMath and TeleTables, evaluate isolated mathematical or tabular reasoning skills. Despite these contributions, current evaluation paradigms exhibit two significant deficiencies. First, they often neglect product-specific documentation—the essential carrier of applied knowledge for actual network equipment—leaving a gap between theoretical understanding and operational reality~\cite{chen2025application}. Second, they focus on ``atomic'' skills while failing to model the complex, multi-step reasoning required in end-to-end industrial processes~\cite{long2025survey}. Although recent works like TeleLogs and OpsEval have begun exploring operational and maintenance tasks~\cite{long2025survey}, they still fall short of capturing the full task pipeline, thereby limiting their ability to holistically measure real-world model performance. Crucially, the lack of product-specific information and realistic workflow contexts in existing benchmarks hinders the transition of telecom-focused LLMs from academic prototypes to production-ready systems~\cite{zou2025telecomgpt,shahid2025large}.

To bridge these gaps, we introduce TeleCom-Bench, a robust benchmark designed for the holistic evaluation of LLMs in the telecommunications domain. By mapping the hierarchy from factual recall to complex problem-solving, TeleCom-Bench provides a definitive measure of an LLM's utility as a specialized agent in the modern telecommunications landscape. Specifically, our framework is structured around two synergistic pillars:
\begin{itemize}[leftmargin=*]
    \item \textbf{Multi-dimensional Knowledge Comprehension}: Beyond 3GPP standards and academic literature, we uniquely incorporate official product manuals and technical patents. We employ a dual-synthesis strategy—combining knowledge graph-driven structural modeling~\cite{xu2025harnessing,gao2025frag} with distillation-driven conceptual extraction~\cite{song2025injecting}—to ensure a granular assessment of both theoretical principles and equipment-specific operational logic~\cite{knollmeyer2025document}.
    \item \textbf{End-to-End Knowledge Application}: We transform real-world trajectories from live operational networks, such as network load optimization~\cite{10829820,11358889} and fault root-cause analysis~\cite{10.1145/3701716.3715225,10979844}, into rigorous evaluation tasks. These scenarios go beyond simple QA, requiring the model to navigate technical constraints, perform causal reasoning, and generate actionable engineering solutions~\cite{panek2025taia}.
\end{itemize}

In summary, the key contributions of this work are as follows:

\begin{itemize} 
\item \textbf{Comprehensive Benchmarking}: To the best of our knowledge, we propose the first systematic telecommunications benchmark that spans both \textbf{foundational knowledge comprehension} and \textbf{complex knowledge application}. By evaluating models across this hierarchy—from the mastery of technical standards to the execution of multi-step industrial workflows, we provide a standardized and multidimensional metric for assessing LLM proficiency in specialized telecom environments~\cite{zhou2024large,long2025survey}.

\item \textbf{Industrial-Oriented Application}: We bridge the gap between theoretical knowledge and practical engineering via focusing on high-level knowledge application in real-world industrial scenarios. By uniquely integrating proprietary equipment manuals with complex telecom service workflows~\cite{chen2025application}, we formalize end-to-end task chains, such as network optimization and fault diagnosis~\cite{10829820,10.1145/3701716.3715225}. This enables a rigorous assessment of an LLM's operational readiness and its effectiveness as a decision-making agent within authentic telecommunications production environments~\cite{zhang2025toward}.

\item \textbf{Empirical Insight \& Validation}: We conduct an extensive comparative analysis of both leading open-source and proprietary LLMs within our framework. By systematically diagnosing failure modes and identifying performance bottlenecks, we provide actionable strategies for model optimization and offer a rigorous roadmap for researchers and engineers in the telecommunications community~\cite{yang2026frontiers,khoramnejad2025generative}.
\end{itemize}

\section{Methodology}
\subsection{Construction of Knowledge Comprehension Evaluation Dataset}
\begin{figure*}[htp]
\centering
\includegraphics[width=\textwidth]{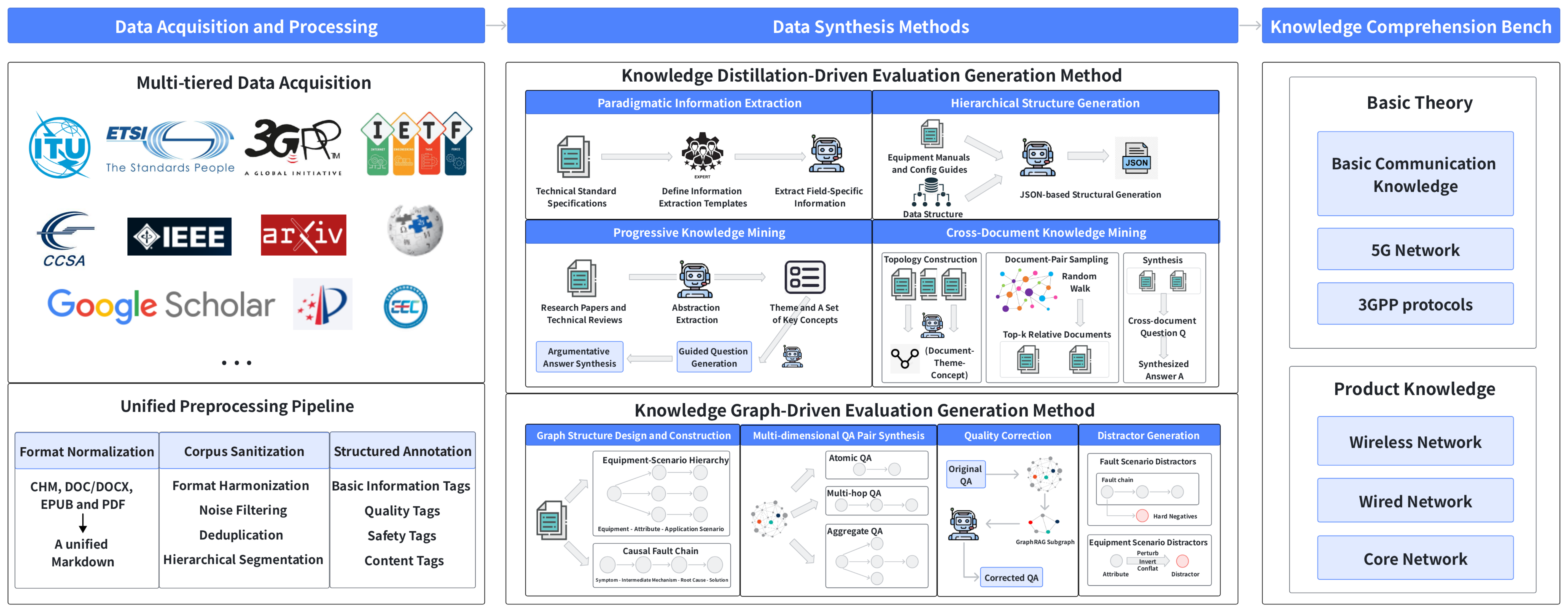}
\caption{Knowledge Comprehension Pipeline: From Acquisition to Evaluation Generation.This pipeline covers three stages and supports the construction of domain-specific knowledge comprehension evaluation datasets with high validity and diversity.}
\label{fig:Knowledge Comprehension}
\end{figure*}

To systematically evaluate the model's knowledge comprehension capability, we first design a complete and scalable knowledge comprehension pipeline that spans the entire process from raw knowledge acquisition to high-quality evaluation sample generation, the detailed framework of which is illustrated in Figure~\ref{fig:Knowledge Comprehension}. 
\subsubsection{Data Acquisition and Processing}
% Constructing a robust knowledge comprehension benchmark for telecommunications necessitates a systematic methodology for data acquisition and curation that spans foundational theory to real-world operational expertise. This section delineates our multi-source data collection framework and the subsequent preprocessing pipeline designed to transform heterogeneous raw materials into a high-quality, structured evaluation corpus.
We transform 1.52TB of heterogeneous raw materials into a high-quality, structured evaluation corpus comprising 17,442 knowledge comprehension evaluation samples through a systematic multi-source data collection framework and a comprehensive preprocessing pipeline \cite{liu2025datasets}.

\textbf{Multi-tiered Data Acquisition}  To ensure comprehensive coverage of both theoretical underpinnings and engineering practice, we aggregate data from two complementary dimensions. Fundamental Knowledge is derived from four authoritative tiers: (i) standardization documents, including 3GPP specifications, IETF RFCs, ITU-T recommendations, and IEEE standards, which provide protocol-level precision \cite{lin20253gpp,lin2025bridge,cox2025introduction,zheng2025validating,garcia2025rfc}; (ii) academic resources comprising IEEE/ACM publications, domain-specific textbooks, and patent disclosures; (iii) curated repositories such as engineering encyclopedias; and (iv) standardized examination banks involving certified professional questions that offer pedagogically validated difficulty gradations.
Complementing this, Product Knowledge spans the entire equipment lifecycle: (i) basic business information, covering feature documents, base protocols, user manuals, and counter description guides; (ii) expert assets, such as systematic business knowledge, algorithm specifications, tool manuals, and diagnostic checklists; (iii) field operational cases documenting real-world problem symptoms and resolutions; and (iv) live service data from intelligent customer support scenarios. This integration bridges the gap between abstract theory and decision-centric industrial reality.

\textbf{Unified Preprocessing Pipeline}  Raw materials undergo a three-stage preprocessing pipeline to ensure linguistic consistency, factual reliability, and evaluation readiness:

\textbf{(1) Format Normalization} Heterogeneous source formats, including CHM, DOC/DOCX, EPUB, and PDF, are converted into a unified Markdown representation. PDFs undergo layout-aware extraction using vision-augmented OCR to handle diagram-rich protocol specifications \cite{nassar2025smoldoclingultracompactvisionlanguagemodel,luo2024layoutllm}. Simultaneously, multimodal elements such as network topology diagrams are processed via Multimodal LLMs to generate descriptive alt-text or preserved as referenced assets with structured captions.

\textbf{(2) Corpus Sanitization}  A multi-phase protocol ensures integrity: (i) Format Harmonization standardizes mathematical notations and units; (ii) Noise Filtering removes non-informative boilerplate like headers and copyright notices; (iii) Deduplication operates at the document level via MD5 and the semantic level using embedding-based clustering \cite{zhou2025megamath,huang2025opencoder,shi2025pre}; and (iv) Hierarchical Segmentation partitions documents into contextually coherent chunks, ranging from protocol clauses to paragraphs, using semantic boundary detection.

\textbf{(3) Structured Annotation}  Segments are enriched with a multi-dimensional metadata system: (i) Basic Information Tags encode attributes like token count and language; (ii) Quality Tags capture metrics including Perplexity, Type-Token Ratio, and LLM-based assessments; (iii) Safety Tags identify toxicity risks; and (iv) Content Tags provide granular categorization of domain topics, licenses, and manual expert annotations. This robust tagging scheme supports stratified benchmark construction and fine-grained model diagnosis.

\subsubsection{Knowledge Distillation-Driven Evaluation Generation Method}
To transform unstructured domain knowledge into structured evaluation tasks, we propose a Task-Adaptive LLM Knowledge Distillation Method. By dynamically tailoring prompt strategies to the intrinsic characteristics of source documents such as structure and knowledge density, this approach enables the scalable and automated synthesis of high-quality evaluation data \cite{wang2023self,wu2024language,zhang2026instruction}.

\textbf{(1) Technical Standard Specifications: Paradigmatic Information Extraction}
Technical standards (e.g., IETF RFCs, 3GPP specifications) are characterized by rigorous language and precise definitions, making them ideal for evaluating factual retrieval and complex instruction-following. We employ a Paradigm-Constrained Extraction method. First, domain experts define information extraction templates (paradigms) $P$ consisting of categories such as protocol names, operating frequency bands, and signaling procedures. Subsequently, the document $D$ and template $P$ are fed into the LLM with instructions to strictly extract field-specific information. By converting open-domain generation into closed-domain slot-filling, we effectively suppress "hallucinations." The resulting $(Document, Paradigm, Result)$ triplets serve as ground truth for evaluating information positioning and fidelity in professional contexts.

\textbf{(2) Equipment Manuals and Config Guides: Hierarchical Structure Generation}
Manuals and configuration guides describe the hierarchical dependencies between system components and parameters. To evaluate the model's ability to parse complex architectures, we utilize a JSON-based Structural Generation method \cite{dong2025xgrammar}. Documents are partitioned into logical blocks and processed to identify entities and attributes. The LLM is prompted to reorganize the content into multi-layer nested JSON objects reflecting the actual system logic. Requiring the model to generate JSON necessitates an explicit understanding of the "part-whole" relationships within the documentation. This dataset assesses performance in knowledge graph construction and complex query answering.

\textbf{(3) Research Papers and Technical Reviews: Progressive Knowledge Mining}  Academic papers are characterized by high knowledge density, necessitating deep cognitive processing. To address this, we propose a Progressive Prompt-Chain Mining workflow composed of three interactive stages. The process initiates with abstraction, where the model performs a preliminary analysis to distill the central theme $T$ and a set of key concepts $C = \{c_1, c_2, \dots, c_n\}$. Building upon $C$, the guided question generation stage yields complex questions $Q$ that demand higher-order cognitive operations---such as analysis, comparison, and evaluation---thereby transcending simple factual recall \cite{wei2022chain,yao2023tree,gao2023prompt}. Finally, in the argumentative answer synthesis stage, the model adopts the persona of a domain expert to construct an answer $A$ strictly grounded in the source text, providing a cohesive logical reasoning chain supported by specific cited evidence.

\textbf{(4) Heterogeneous Technical Documents: Cross-Document Knowledge Mining}  Resolving engineering problems often necessitates synthesizing information from multiple sources. To ensure our dataset captures this complexity, we introduce a Cross-Document Knowledge Topology Synthesis method. Initially, we preprocess the corpus to extract ``Document-Theme-Concept'' associations, constructing a lightweight topology graph $\mathcal{G}$ where edges represent semantic correlations. Subsequently, a constrained random walk on $\mathcal{G}$ yields a concept subset $C_{walk}$, which guides the retrieval of document pairs $(D_\alpha, D_\beta)$ that exhibit high semantic overlap yet offer complementary perspectives, such as theoretical principles versus engineering practice. Finally, an LLM is tasked with generating a cross-document question $Q_{cross}$ and a comprehensive answer $A_{cross}$, necessitating the explicit fusion of information from both $D_\alpha$ and $D_\beta$.

\subsubsection{Knowledge Graph-Driven Evaluation Generation Method}
We propose a Knowledge Graph-driven method to construct a telecommunications evaluation benchmark. By explicitly modeling global knowledge correlations via interpretable graph topologies, this approach mitigates the limitations of probabilistic prompt synthesis, such as cross-document fragmentation, hallucination, and opacity \cite{chen2025graphgenenhancingsupervisedfinetuning,cheng2025neural,dai2025large}.

\textbf{(1) Scenario-Based Graph Structure Design and Construction}  To rigorously represent professional knowledge, we transform unstructured raw corpora into a structured entity-relation graph governed by a predefined ontology. For general scenarios, we employ predefined entity types (e.g., \texttt{NetworkFunction}, \texttt{Interface}, \texttt{KPI}) and explicit semantic edges (e.g., \texttt{[causes]}, \texttt{[flow\_pair]}, \texttt{[impacts\_kpi]}) to construct a comprehensive undirected connected knowledge graph. Specifically, tailored to equipment and fault scenarios, we engineered two specialized topological structures, the Equipment-Scenario Hierarchy Tree and the Causal Fault Chain, to capture distinct logical dependencies:

\textbf{Equipment-Scenario Hierarchy}  For equipment corpora, we construct a hierarchical tree structure of "Equipment - Attribute - Application Scenario". By modeling granular attributes such as frequency band, channel number, and power, we explicitly derive the mapping to application scenarios like high-speed rail coverage and urban hotspots.

\textbf{Causal Fault Chain}  For fault-related corpora, we construct a complete causal chain comprising "Symptom - Intermediate Mechanism - Root Cause - Solution". Furthermore, we mine analogous trigger sources across disjoint chains to maximize the associative density of the knowledge graph.

\textbf{(2) Multi-dimensional QA Pair Synthesis Based on Graph Traversal}  Leveraging the global KG, we employ graph traversal algorithms to sample disjoint connected subgraphs as seed units and synthesize QA pairs across three distinct granularities: Atomic QA (Factual Retrieval), which utilizes DFS/BFS to traverse individual nodes or edges and extract entity-relation triples for specific factual knowledge; Multi-hop QA (Sequential Reasoning), which constructs subgraphs via BFS to establish sequential reasoning chains, testing multi-step deduction capabilities \cite{tan2025paths}; and Aggregate QA (Holistic Synthesis), which targets larger-scale semantic clusters, requiring the model to synthesize information from the entire subgraph to generate comprehensive questions. This mechanism explicitly tracks node and edge usage to guarantee non-overlapping sampling, fundamentally eliminating dataset redundancy.

\textbf{(3) Quality Correction Based on GraphRAG Subgraphs}  To mitigate generation artifacts such as hallucinations, we introduce a Graph-Augmented Retrieval correction module \cite{min2025towards,peng2025graph,zhang2025survey}. By retrieving local context subgraphs via neighbor expansion, this module refines QA pairs to ensure contextual completeness through supplemented constraints, performs noise reduction by filtering irrelevant information, and guarantees factual faithfulness by aligning generated details with verified KG paths.

\textbf{(4) Distractor Generation Based on Subgraph Differences}
To enhance the discriminative power of the evaluation, we exploit graph structural properties to generate adversarial distractors, replacing simplistic random negation \cite{rani2025automated}.

\textbf{Fault Scenario Distractors}  Selecting a valid fault chain as the anchor, we traverse the graph to identify "hard negatives"---nodes from parallel fault mechanisms or conceptually similar but causally distinct trigger sources. This ensures distractors are technically plausible but logically invalid.

\textbf{Equipment Scenario Distractors}  Based on the Equipment-Scenario Hierarchy, we design options targeting subtle attribute variations (e.g., differing max remote distances). Strong distractors are constructed by perturbing attribute values, inverting process sequences, or conflating functional attributions of distinct equipment entities.
Finally, we employ graph-augmented LLMs to validate the plausibility and difficulty of distractors, guaranteeing that the overall evaluation benchmark achieves high discrimination and rigorous quality standards.

\subsection{Construction of Knowledge Application Evaluation Dataset}
\begin{figure*}[htp]
\centering
\includegraphics[width=\textwidth]{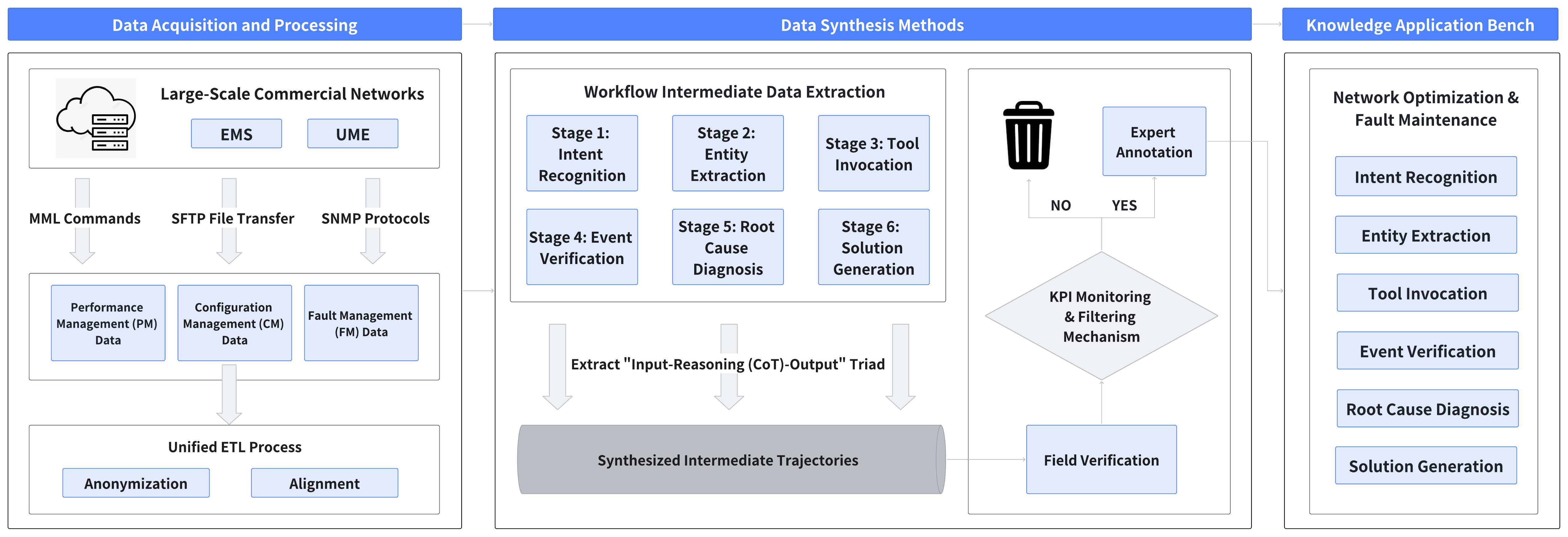}
\caption{Knowledge Application Pipeline: From Acquisition to Evaluation Generation.This pipeline illustrates the full workflow from multi-source network data acquisition, synthesis, to knowledge application benchmark construction.}
\label{fig:Knowledge Application}
\end{figure*}

Unlike knowledge comprehension tasks that focus on static information retrieval, knowledge application requires models to interact with dynamic environments. We propose a robust construction pipeline that synergizes real-world telemetry with agent-based workflow mining \cite{zhang2024survey}, whose overall framework is illustrated in Figure~\ref{fig:Knowledge Application}.

\subsubsection{Data Acquisition and Preprocessing}

To ensure the benchmark reflects the complexity of real-world networks, we collect raw data from large-scale commercial networks via Network Element Management Systems and Unified Management Environments. The data acquisition utilizes hybrid protocols including MML commands, SFTP file transfer, and SNMP.
The dataset encompasses three core dimensions: Performance Management data captures dynamic network KPIs via 15-minute time-series matrices to quantify fluctuating service quality; Configuration Management data provides static snapshots of topology and engineering parameters that define the feasible optimization state space; and Fault Management data comprises event-driven logs of system anomalies, offering critical contextual clues for root cause analysis.
We implement a privacy-preserving ETL pipeline to clean the raw data \cite{majeed2020anonymization,mokale2024data,manda2023privacy}. Sensitive identifiers such as IMSI and GPS coordinates are anonymized using irreversible hashing. Subsequently, asynchronous PM series and FM logs are temporally aligned with CM snapshots to construct coherent Network State Observations.

\subsubsection{Data Synthesis Methods}
Traditional simulation-based evaluation often suffers from the "reality gap." We address this by employing an Agent-in-the-Loop approach using live network data as the environment. We deploy domain-specific agents to solve actual network issues and utilize a Trajectory Mining technique to extract structured evaluation samples from their execution traces.
We decompose the complex optimization workflow into atomic evaluation tasks, capturing the Input Context, Chain-of-Thought, and Action triad at each stage:

\begin{itemize}[leftmargin=*]

\item \textbf{Intent Recognition and Entity Extraction}  We record the transformation process from ambiguous natural language instructions to precise, structured intent schemas, alongside the grounding of topological data into specific Network Elements.

\item \textbf{Tool Invocation and Event Verification}  We log not only the API calls but also the strategic decision-making process—capturing why specific diagnostic tools were selected based on the current context—and the logical verification steps that filter raw logs into confirmed anomalies.

\item \textbf{Root Cause Diagnosis and Solution Generation}  We preserve the complete Chain-of-Thought that bridges observed symptoms to root causes, as well as the multi-objective trade-off analysis that underpins the final optimization solutions \cite{wei2022chain}.
\end{itemize}

\textbf{Closed-Loop Field Verification}  Generated optimization actions are provisioned into the live commercial network under strict safety guardrails. We measure the causal impact by comparing KPIs such as user perceived rate before and after execution. Only trajectories that achieve statistically significant gains without triggering Performance Degradation Alarms are retained. This guarantees that the "Ground Truth" is physically verifiable.

\textbf{Expert Review}  Verified trajectories undergo a final audit by senior engineers. Experts refine the semantic description of the reasoning steps and correct any subtle logical flaws, ensuring the dataset aligns with high-standard engineering protocols.

\section{Benchmark Design}
\label{sec:dataset}

% To systematically evaluate the capabilities of LLMs and domain-specialized agents in telecommunications, we constructed the \textsc{TeleCommuBench} dataset, a comprehensive benchmark encompassing both foundational Knowledge Comprehension and practical problem-solving skills. The dataset is organized into two primary categories, further divided into multiple subcategories and specific task types, with detailed statistics summarized in Table~\ref{tab:dataset_stats}. As illustrated in Figure~\ref{fig:dataset_structure}, \textsc{TeleCommuBench} comprises a total of \textbf{22,409} meticulously curated evaluation samples.

To systematically evaluate telecommunications LLMs and agents, we present TeleCom-Bench, a comprehensive benchmark comprising 12 specialized subsets and 22,678 curated samples (see Table~\ref{tab:task_distribution} for details). The benchmark is structured into two synergistic categories: Knowledge Comprehension, which assesses static mastery of fundamental theories and product specifications; and Knowledge Application, which evaluates dynamic problem-solving within authentic end-to-end workflows like network optimization and fault maintenance. Distinguished by high industrial fidelity, TeleCom-Bench is constructed from real-world configurations, telemetry, and operational tickets, filling critical gaps in equipment product knowledge and establishing a robust framework for validating LLMs' capabilities in closed-loop scenarios.

\begin{table*}[t]
\centering
\caption{Hierarchical Composition and Scale of TeleCom-Bench: Task Taxonomy, Sample Distribution, Synthesis Methodology}
\label{tab:task_distribution}
\begin{tabular}{lllllr}
\toprule
\textbf{Level 1} & \textbf{Level 2} & \textbf{Construction Method} & \textbf{Task Type} & \textbf{Task Name} & \textbf{Count} \\
\midrule
\multirow{6}{*}{Comprehension} & \multirow{3}{*}{Basic Theory} & \multirow{3}{*}{Knowledge Distillation} & \multirow{3}{*}{Multiple-Select Questions} & Basic Knowledge & 2,662 \\
& & & & 5G Network & 2,564 \\
& & & & 3GPP Protocols & 4,043 \\
\cmidrule(lr){2-6}
& \multirow{3}{*}{Product Knowledge} & \multirow{3}{*}{Knowledge Graph} & \multirow{3}{*}{Subjective QA} & Wireless Network & 3,725 \\
& & & & Wired Network & 3,488 \\
& & & & Core Network & 960 \\
\midrule
\multirow{6}{*}{Application} & \multirow{6}{*}{\shortstack[l]{Network Optimization \&\\ Fault Maintenance}} & \multirow{6}{*}{\shortstack[l]{Agent-Driven\\ Trajectory Synthesis}} & \multirow{6}{*}{Structured QA} & Intent Recognition & 2,174 \\
& & & & Entity Extraction & 365 \\
& & & & Tool Invocation & 585 \\
& & & & Event Verification & 146 \\
& & & & Root Cause Diagnosis & 983 \\
& & & & Solution Generation & 983 \\
\midrule
\multicolumn{5}{l}{\textbf{Total}} & \textbf{22,678} \\
\bottomrule
\end{tabular}
\end{table*}

\subsection{Knowledge Comprehension Evaluation}
\label{sec:Knowledge Comprehension Evaluation}
% The Knowledge Comprehension category focuses on evaluating whether the model has mastered the full spectrum of communication domain knowledge. It is assessed in two parts: General Communication Knowledge, which examines the understanding of concepts and mastery of general knowledge (e.g., 5G, 3GPP protocols, network architecture); and Communication Product Knowledge, which fills the gap in current benchmarks regarding mainstream vendor product information covering Wired, Wireless, and Core networks, specifically testing the mastery of business concepts and product features.
The Knowledge Comprehension category evaluates LLMs' mastery of communication domain knowledge. It comprises two parts: Communication Basic Theory, assessing theoretical foundations like 5G architecture and 3GPP protocols; and Communication Product Knowledge, which bridges benchmark gaps by testing product concepts and workflows across Wired, Wireless, and Core networks.

\subsubsection{Communication Basic Theory}  The Communication Basic Theory evaluation comprises three dimensions: General Fundamentals, which covers information theory, signal processing, wireless propagation, and network principles based on the OSI and TCP/IP models; 5G Network Architecture, which assesses system topology including SA/NSA modes, Core SBA, and RAN functional splits; and 3GPP Protocols, which examines the full stack from Release 15 to 17, spanning RRC, MAC, PDCP, and NAS signaling.

\subsubsection{Communication Product Knowledge}
% This part of the evaluation set covers mainstream product information across Wired, Wireless, and Core networks, filling the gap in current communication domain QA benchmarks regarding mainstream vendor product information. It mainly examines the model's mastery of business concepts and product features. We have extensively collected high-value internal and practical data, including Basic Business Information (business feature documents, basic protocols, user manuals, operation manuals, and counter description documents), Expert Knowledge (business thematic documents summarized by experts, systematic business knowledge documents, algorithm documents, tool manuals, troubleshooting guides, and checklists), Real-world Cases (actual cases from on-site network systems, including problem phenomena, troubleshooting methods, and recommendation summaries), and Live Data (data backflow from intelligent customer service across various business scenarios).

This part of the evaluation set addresses the scarcity of vendor-specific product knowledge in existing benchmarks by covering mainstream products and equipment information across Wired, Wireless, and Core networks. It rigorously evaluates LLMs' grasp of product concepts, functional features, and operational workflows, bridging the gap between theoretical knowledge and practical product application.

\textbf{Core Network Product Distribution}  The Core Network evaluation targets virtualization platforms and critical network functions such as PCF, UDM, and AMF. It assesses three dimensions comprising Product Concepts, Deployment Workflows, and Operations and Maintenance. Furthermore, the scope extends to essential infrastructure and specialized modules including Distributed Storage, NFV Orchestrators, and Charging Gateways.
% The Core Network evaluation set covers mainstream products such as TECS\_OpenStack, RCP, USPP, and uAMC. The questions are categorized into Product Concept Q\&A, Product Workflows, and Product Operations \& Maintenance Information. Other covered products include TECS\_CloveStorage, UME\_R50, TECS\_Director, CG, CNIA (xDR/MDAF), VNFM, xGW, GSO, 5GMC, and Rack\_Server.

\textbf{Wireless Network Product Distribution}  The Wireless Network segment focuses on 5G and 4G access solutions, centering on critical hardware including Base Station components like BBU, RRU, and AAU as well as Antenna Systems. Key evaluation criteria encompass hardware architecture analysis, RF indicator interpretation, parameter configuration, and optimization strategies for specific scenarios such as high-density urban areas.
% The Wireless Network evaluation set primarily covers general wireless access products for 5G and 4G networks. Key product categories include Base Stations (BBU, RRU, AAU), Antenna Systems, and Indoor Distribution Systems. The evaluation focuses on product hardware architecture, radio frequency indicators, cell parameter configuration, and typical application scenarios (e.g., high-speed rail, high-density urban areas).

\textbf{Wired Network Product Distribution}  Anchored by SPN and PTN, the Wired Network category covers key technologies including MPLS, SRv6, and routing protocols like BGP, IS-IS, and OSPF. The evaluation extends to Router and Switch products with a focus on reliability and ring network protocols, alongside Optical Transport Network technologies. It also encompasses broader infrastructure such as IPRAN and vBRAS.
% The distribution across key product categories is as follows: SPN \& PTN (Slice Packet Network \& Packet Transport Network) constitutes the largest category, covering MPLS technologies (TP/L3VPN/TE), Packet Transport (VPWS/VPLS/PWE3), Routing Protocols (BGP/IS-IS/OSPF), Label Technologies (SRv6/SR-MPLS), and FlexE. Router products cover routers, routing protocols, policy routing, and reliability protocols (BFD/NSR). Switch products cover switches, VLAN/MAC management, and ring network protocols (ERPS/LACP). The OTN (Optical Transport Network) category focuses on optical transmission technologies, modules, interfaces, and error detection (FEC). Finally, Other Categories include IPRAN, Data Products (vBRAS/vRouter), MSTP, and general wired products.

\subsection{Knowledge Application Evaluetion}
\label{sec:Knowledge Application Evaluetion}

% This section details the construction of the \textsc{XXX} benchmark dataset, focusing on its structure, composition, and evaluation dimensions. A core contribution lies in the design of the \textit{Knowledge Application} category, which assesses a model's capability to solve real-world operational problems within authentic telecommunications workflows.

The Knowledge Application category is designed to assess LLMs' capability to synthesize and apply domain expertise to address complex, end-to-end business scenarios. To simulate real-world workflows, we designed two integrated evaluation pipelines: Network Optimization, covering the full cycle from problem detection to closed-loop optimization, and Fault Maintenance, emphasizing rapid fault localization and recovery. Supporting these scenarios, we decompose typical telecommunications workflows into six sequential competency tasks: Intent Recognition, Entity Extraction, Event Verification, Tool Invocation, Root Cause Diagnosis, and Solution Generation. Dedicated datasets for each task ensure precise evaluation of specific capabilities within these broader workflows.

% The \textit{Knowledge Application} category is designed to evaluate a model's ability to integrate and apply domain knowledge to address practical, end-to-end business scenarios. We deconstruct typical operational workflows in telecommunications into seven sequential competency tasks: \textbf{Intent Recognition}, \textbf{Entity Extraction}, \textbf{Event Verification}, \textbf{Tool Invocation}, \textbf{Root Cause Diagnosis}, and \textbf{Solution Generation}. Evaluation datasets are constructed for each task. Furthermore, to assess holistic performance, we design integrated, end-to-end evaluation pipelines for two critical operational scenarios: \textit{Network Optimization} and \textit{Fault Maintenance}.

\textbf{Intent Recognition}  Given natural language inputs, LLMs must accurately classify the specific business scenario or intent. This task evaluates the model's comprehension of operational terminology and classification accuracy, serving as the critical entry point for subsequent workflows.

\textbf{Entity Extraction}  Building upon intent recognition, this task requires LLMs to precisely extract critical informational entities from unstructured or semi-structured text. This assesses the model's information extraction and structuring capabilities, providing essential inputs for downstream data querying.

\textbf{Event Verification} Based on extracted entities, LLMs must validate network events or states by processing structured data retrieved from management systems via calculation, comparative analysis, and logical reasoning. This focuses on evaluating proficiency in handling structured data and executing rule-based verification.

\textbf{Tool Invocation}  To execute complex tasks, LLMs must orchestrate external tools or APIs by generating correct invocation sequences with appropriate parameter mapping. This assesses capabilities in task decomposition, tool selection, and parameter adaptation necessary for autonomous operations.

\textbf{Root Cause Diagnosis}  LLMs must analyze multi-modal data, including alarms and performance metrics, leveraging domain knowledge to deduce the root cause of identified issues. This deeply evaluates causal reasoning, information fusion, and the application of expert knowledge.

\textbf{Solution Generation}  Upon diagnosis, LLMs must formulate concrete and actionable remediation strategies that directly address the specific root cause. This evaluates the model's solution planning, knowledge transfer, and judgment regarding executability.

Through this targeted task decomposition and integrated pipeline design, the TeleCom-Bench dataset enables a granular and comprehensive assessment of LLMs or AI agents in applying knowledge and solving problems throughout complex, dynamic, real-world telecommunications operational workflows.

\section{Experiments and Analysis}
\label{sec:experiments}
% In this section, we present a systematic evaluation of representative Large Language Models on TeleCom-Bench. Our analysis aims to dissect the discrepancy between theoretical proficiency and operational viability, specifically investigating the "Execution Wall" where linguistic competence fails to translate into actionable engineering solutions.

\begin{table*}[t]
\centering
\caption{Main Results on TeleCom-Bench. The table reveals a critical "Action Gap." While large models like Qwen3-235B excel at diagnosing root causes (71.49\%), they fail catastrophically when tasked with generating executable solutions (4.67\%). DeepSeek-V3.2 demonstrates superior tool invocation capabilities but similarly struggles with final solution generation.}
\label{tab:main_results}
\resizebox{\textwidth}{!}{%
\begin{tabular}{@{}llcccccccc@{}}
\toprule
\textbf{Category} & \textbf{Task} & \textbf{Qwen3-32B} & \textbf{Qwen3-235B} & \textbf{DeepSeek-V3.2} & \textbf{Gemini2.5} & \textbf{Grok4.1} & \textbf{GLM4.7} & \textbf{Doubao-pro} & \textbf{Kimi K2} \\ \midrule
{Comprehension} & Basic Theory & 61.01 & 69.04 & 67.12 & 63.88 & \textbf{70.97} & 51.82 & 70.14 & 52.75\\
 & Wired Network & 59.41 & 60.88 & 57.34 & 54.68 & \textbf{62.60} & 60.90& 58.78 & 60.99 \\
 & Wireless Network & \textbf{73.04} & 70.80 & 45.89 & 63.64 & 52.58 & 22.63 & 61.99 & 64.81 \\
 & Core Network & 60.22 & 62.13 & 62.34 & 59.28 & \textbf{66.79} & 47.68 & 64.46 & 56.79 \\ \midrule
{Application} & Intent Recognition & 93.13 & \textbf{94.52} & 92.85 & 92.94 & 93.85 & 92.69 & 93.43 & 93.49 \\
 & Entity Extraction & 95.74 & \textbf{99.72} & \textbf{99.72} & \textbf{99.72} & \textbf{99.72} & 81.25 & \textbf{99.72} & \textbf{99.72} \\
 & Event Verification & 59.00 & 72.72 & 67.35 & 71.92 & \textbf{81.85} & 12.95 & 80.48 & 52.92 \\
 & Tool Invocation & 84.71 & 45.54 & \textbf{94.06} & 90.80 & 93.20 & 56.50 & 87.30 & 84.50 \\
 & Root Cause Diagnosis & 60.92 & \textbf{71.49} & 63.00 & 49.28 & 48.60 & 26.63& 61.33& 57.85 \\
 & Solution Generation & 15.02 & 4.67 & 5.61 & 22.42 & 14.58& 9.64& \textbf{30.72} & 8.45 \\ \bottomrule
\end{tabular}%
}
\end{table*}
In this section, we present a systematic evaluation of representative Large Language Models on TeleCom-Bench. Our analysis quantifies the discrepancy between theoretical proficiency and operational viability in telecommunications engineering, with particular focus on the Execution Wall—defined operationally as the performance gap exceeding 50 percentage points between diagnostic reasoning and executable solution generation within the same fault-handling workflow.

\subsection{Experimental Setup}
% We evaluated a diverse spectrum of models, ranging from efficient open-weights models like \textbf{Qwen3-32B} to massive scale models like \textbf{Qwen3-235B} (Dense) and \textbf{DeepSeek-V3.2} (MoE, $\approx$671B parameters). We also included proprietary APIs such as \textbf{Gemini 2.5}, \textbf{Grok 4.1}, and \textbf{GLM-4.7} for a comprehensive comparison.

% Evaluation metrics were tailored to task specificity: \textit{Macro-F1} for multiple-choice theoretical questions, \textit{Exact Match Accuracy} for structured extraction tasks, and a calibrated \textit{LLM-as-a-Judge} mechanism for open tasks (e.g., Core Network), ensuring alignment with expert-annotated ground truth.

\subsection{Analysis of Knowledge Comprehension}

\textbf{Models}  We evaluated eight representative large language models namely Qwen3-32B, Qwen3-235B-A22B, DeepSeek-V3.2, Gemini 2.5, Grok 4.1, GLM-4.7, Doubao-pro, and Kimi K2. To ensure fair and comparable assessment across diverse architectures, all models were evaluated under identical inference conditions with temperature set to 0.7, reasoning mode enabled to support multi-step deliberation, and three independent sampling trials performed per instance with the final prediction determined by majority voting.
\textbf{Evaluation Metrics}  We adopt Macro-F1 for multiple-response questions to address class imbalance in multi-label classification, Exact Match for structured QA tasks to enforce strict format and content fidelity, and a tri-expert LLM-as-a-Judge voting mechanism scoring open-ended responses on a 5-point Likert scale with Krippendorff's $\alpha$=0.82 inter-annotator agreement against human experts.

Results in the Knowledge Comprehension category as shown in Table~\ref{tab:main_results} indicate that state-of-the-art LLMs achieve moderate proficiency (60\%--70\%) in foundational telecom theory but exhibit substantial variance in product-specific knowledge. Notably, Qwen3-32B outperforms its larger counterpart Qwen3-235B on \textit{Wireless Network} (73.04\% vs. 70.80\%), while GLM-4.7 achieves only 22.63\%. This non-monotonic scaling suggests that pre-training data distribution—particularly coverage of vendor-specific documentation—plays a more decisive role than parameter count alone in vertical domains. The consistent sub-70\% performance across all models confirms that proprietary operational knowledge remains a persistent blind spot for publicly available LLMs.

% The results in the \textit{Knowledge Comprehension} (Table \ref{tab:main_results}) indicate that generalist models have achieved a "passing grade" but lack mastery. Most top-tier models (Grok4.1, Doubao-pro, Qwen3) cluster around 60\%--70\% in \textit{Basic Theory}. 

% However, a significant variance is observed in domain-specific product knowledge, particularly in \textit{Wireless Network}. Qwen3-32B achieves a remarkable 73.04\%, significantly outperforming GLM4.7 (22.63\%) and even surpassing its larger counterpart Qwen3-235B (70.80\%). This suggests that for highly specialized vertical domains, the distribution of pre-training data is often more critical than pure parameter scale. The general struggle to exceed 70\% across all product categories confirms that proprietary vendor knowledge remains a cognitive blind spot for public models.

\subsection{Analysis of Knowledge Application}
% The most critical insights of TeleCom-Bench lie in the \textit{Knowledge Application} results expose a stark "Execution Wall," revealing that operational capability does not scale linearly with theoretical knowledge or model size.
% As illustrated in Figure \ref{fig:detailed_subplots}, the architectural disparity between MoE and dense models translates into distinct "capability profiles" on our radar charts. While DeepSeek-V3.2 and Qwen3-235B demonstrate robust coverage in diagnostic tasks like Root Cause Analysis and Event Verification, they both exhibit a sharp inward collapse in the Solution Generation sector.

\begin{figure*}[t]
    \centering
    \includegraphics[height=6.3cm,keepaspectratio]{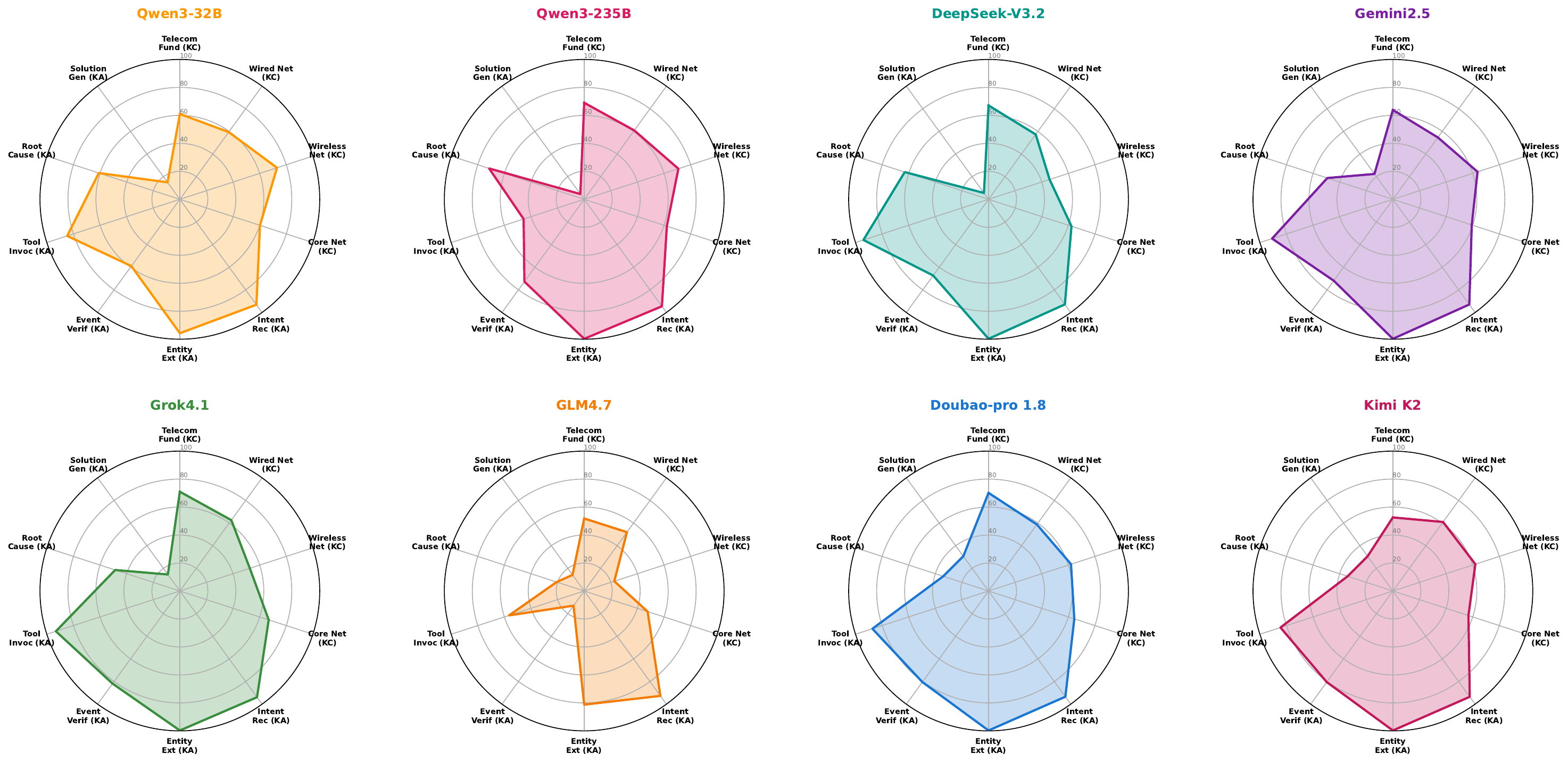}
    \caption{Performance radar charts of diverse LLMs across Telecom-specific capabilities. The benchmarks are categorized into Knowledge Comprehension (KC) and Knowledge Application (KA). Note the consistent "structural deficiency" across all models in Solution Generation, contrasted with relatively high scores in Root Cause Diagnosis and Entity Extraction.}
    \label{fig:detailed_subplots}
\end{figure*}

The Knowledge Application results expose a systematic Execution Wall: operational capability does not scale linearly with theoretical knowledge or model size. As visualized in Figure~\ref{fig:detailed_subplots}, all models exhibit a structural deficiency in \textit{Solution Generation}, despite strong performance in upstream tasks.

% --- FIGURE: DETAILED SUBPLOTS ---

\subsubsection{Saturated Linguistic Interface}
In early workflow stages (Intent Recognition and Entity Extraction), performance approaches saturation ($>$92\% and $>$95\%, respectively). This confirms that the linguistic interface problem is largely solved: models reliably parse expert instructions and extract structured parameters. However, this high accuracy creates a reliability illusion that dissipates in downstream execution tasks.

\subsubsection{Architecture-Dependent Tooling Capability}
A pronounced architectural divergence emerges in \textit{Tool Invocation}. DeepSeek-V3.2 achieves 94.06\%, significantly outperforming the dense Qwen3-235B (45.54\%). Pearson correlation analysis reveals a strong positive association between MoE architecture and tool-calling proficiency ($r$=0.87, $p$<0.01), suggesting that sparse activation combined with targeted post-training for function calling better supports dynamic tool orchestration in telecom automation.

\subsubsection{The Diagnosis-Action Paradox}
The most salient finding is the Diagnosis-Action Paradox: models capable of accurate fault diagnosis consistently fail to generate executable remediation plans. Qwen3-235B achieves 71.49\% in Root Cause Diagnosis but collapses to 4.67\% in Solution Generation—a 66.82-point gap defining the Execution Wall. DeepSeek-V3.2 shows a similar pattern (63.00\% $\rightarrow$ 5.61\%). This indicates a fundamental limitation in procedural agency: while models excel at causal reasoning (linking symptoms to causes), they lack the capability for procedural synthesis (converting causes into safety-compliant, executable command sequences).

Even the top-performing Doubao-pro (30.72\% in solution generation) remains far below the 95\% threshold required for autonomous deployment in production networks. This universal performance ceiling underscores that generating executable, safety-constrained maintenance scripts represents the most significant barrier to telecom automation.

% \subsubsection{The Illusion of Linguistic Competence}
% In the initial workflow stages: \textit{Intent Recognition} and \textit{Entity Extraction}, the performance is near-saturated. Virtually all models achieve $>$92\% in intent recognition and $>$95\% in entity extraction. This confirms that the "Linguistic Interface" problem is solved; models can perfectly parse expert instructions and extract structured parameters. However, this creates a deceptive illusion of reliability that dissolves in subsequent steps.

% \subsubsection{The Logic and Tooling Bottleneck}
% As the workflow progresses to \textit{Event Verification} and \textit{Tool Calling}, these tasks highlight a critical insight regarding model architecture. 
% DeepSeek-V3.2, utilizing a massive Mixture-of-Experts architecture ($\approx$671B parameters), achieves a dominant score of \textbf{94.06\%}. In stark contrast, the dense Qwen3-235B model struggles significantly at \textbf{45.54\%}, despite its strong reasoning capabilities elsewhere.
% This disparity indicates that simply scaling up dense parameters does not automatically yield agentic agility. The superior performance of DeepSeek-V3.2 suggests that MoE architectures, likely combined with targeted post-training for function calling, are far more effective for the dynamic tool use required in telecom automation.

\textbf{Summary Insight}  TeleCom-Bench demonstrates that current LLMs function effectively as Information Extractors and competent Diagnosticians, but fail as Field Engineers. Bridging the Execution Wall requires moving beyond general reasoning toward specialized alignment in procedural execution, safety-constrained generation, and tool-grounded interaction.

\section{Related Work}

\textbf{Telecommunications Benchmarks}  The evaluation of Large Language Models in telecommunications has progressed from foundational NLP tasks to complex operational capabilities. Initial research focused on general protocol comprehension, exemplified by SPEC5G~\cite{karim2023spec5g} for document classification and TeleQnA~\cite{maatouk2025teleqna} for concept-based QA. Subsequent studies expanded task diversity and depth: TSpec-LLM~\cite{nikbakht2024tspec} covered broadly across 3GPP standards, ORAN-Bench-13K~\cite{gajjar2025oran} introduced interface code comprehension, and TelecomGPT~\cite{zou2025telecomgpt} incorporated mathematical reasoning.  Comprehensive evaluation suites such as TelcoLM~\cite{barboule2024telcolm} further integrated linguistic analysis (Linguistic~\cite{ahmed2024linguistic}) with domain-specific tasks. 

More specialized efforts have shifted toward operational verticals, such as structured table parsing in TeleTables~\cite{ezzakri2025teletables}, root cause analysis via drive-test logs in TeleLogs~\cite{sana2025reasoning} and OpsEval~\cite{liu2025opseval}, and intent-to-configuration translation in TeleYAML~\cite{ethiraj2025efficient}. 

Collectively, these works establish a stratified assessment hierarchy: a foundational tier based on terminology understanding; a technical tier for protocol and quantitative reasoning; and an application tier focusing on operational value. Despite this progress, three critical limitations persist. First, task fragmentation means evaluations remain isolated without assessing unified knowledge-to-action capabilities. Second, a significant data bias exists as benchmarks rely predominantly on public standards, neglecting product-specific operational knowledge, such as vendor manuals and field-validated procedures that distinguishes authentic expertise from generic literacy. Third, workflow absence results in a lack of end-to-end modeling for multi-step, tool-augmented processes characteristic of real-world network operations.

\textbf{Methodologies in Benchmark Construction.} Contemporary benchmark construction in telecommunications primarily relies on LLM-augmented generation from public standards or structured template synthesis for specific tasks~\cite{maatouk2025teleqna,nikbakht2024tspec,ezzakri2025teletables,colle2025telemath}, which, while scalable, source data almost exclusively from public artifacts and neglect authentic operational environments. This limitation contrasts with emerging practices in adjacent high-stakes domains: CNFinBench reconstructs end-to-end financial agent workflows from certified regulatory corpora and simulates multi-turn adversarial interactions to quantify compliance drift~\cite{ding2025cnfinbench}, while MedBench establishes nationwide evaluation infrastructures with 700K+ tasks refined through multi-stage review by clinicians from 500+ institutions, explicitly aligning assessments with real clinical workflows~\cite{ding2025medbench}. These cross-domain efforts demonstrate that rigorous benchmark construction in risk-sensitive settings demands three methodological pillars—deep expert involvement throughout curation, grounding in authentic operational sequences rather than isolated artifacts, and explicit modeling of tool-mediated multi-step execution—all of which remain absent in current telecommunications benchmarks. Consequently, existing approaches cannot evaluate whether models navigate the proprietary, tool-augmented reality of network operations, creating a critical blind spot in assessing industrial deployment readiness. Our work bridges this gap by integrating vendor-specific documentation and constructing evaluation items entirely from real-world agent workflow traces, thereby establishing the first telecommunications benchmark that satisfies the methodological rigor demonstrated in adjacent high-stakes domains.

\section{Conclusion}
% In this paper, we present XXX, a systematic benchmark for evaluating large language models (LLMs) in the telecommunications domain, built from real-world product documentation and real workflows using knowledge graph–based, agent-driven, and model-enhanced data construction methods. Experiments show that while LLMs excel on general-knowledge tasks, they struggle with telecom-specific product knowledge and operational workflows. XXX validates the potential of LLMs in this domain and highlights critical directions for future model development.

% In this paper, we introduce TeleCom-Bench, the first benchmark for evaluating large language models in industrial telecommunications. Spanning 12 evaluation sets with 22,678 curated samples, it establishes a two-tier framework: (1) Multi-dimensional Knowledge Comprehension, integrating communication theory, 3GPP protocols, and vendor-specific product knowledge across wired, core, and wireless networks; and (2) End-to-End Knowledge Application, formalizing six sequential tasks derived from authentic agent trajectories in live network operations. Our evaluation reveals a fundamental gap between diagnostic reasoning and executable action generation, demonstrating that domain alignment through proprietary documentation outweighs model scale in industrial effectiveness. TeleCom-Bench provides standardized metrics to pinpoint capability gaps and accelerates the transition of LLMs from academic prototypes to production-ready agents in real-world telecommunications networks.

In this paper, we introduce TeleCom-Bench, a benchmark for evaluating large language models in industrial telecommunications. It covers 12 evaluation sets, 22,678 samples and measures two capability tiers: (1) Multi-dimensional Knowledge Comprehension over communication theory, 3GPP protocols, and vendor-specific product knowledge; and (2) End-to-End Knowledge Application over six sequential tasks grounded in real network-operation trajectories. Results highlight a persistent gap between accurate diagnosis and executable action generation, motivating future work on domain alignment and safety-constrained procedural execution.
\clearpage

% Start references on a fresh page so the Conclusion can use the remaining
% space in the current two-column page.
\clearpage

%% the bibliography file(s).
\bibliographystyle{ACM-Reference-Format}
\bibliography{sample-base,software}

\appendix
\section{Appendix}

\subsection{A.1 Question Format Specifications in \textsc{TeleCom-Bench}}
\begin{figure*}[h]
\centering
\includegraphics[width=1.0\linewidth]{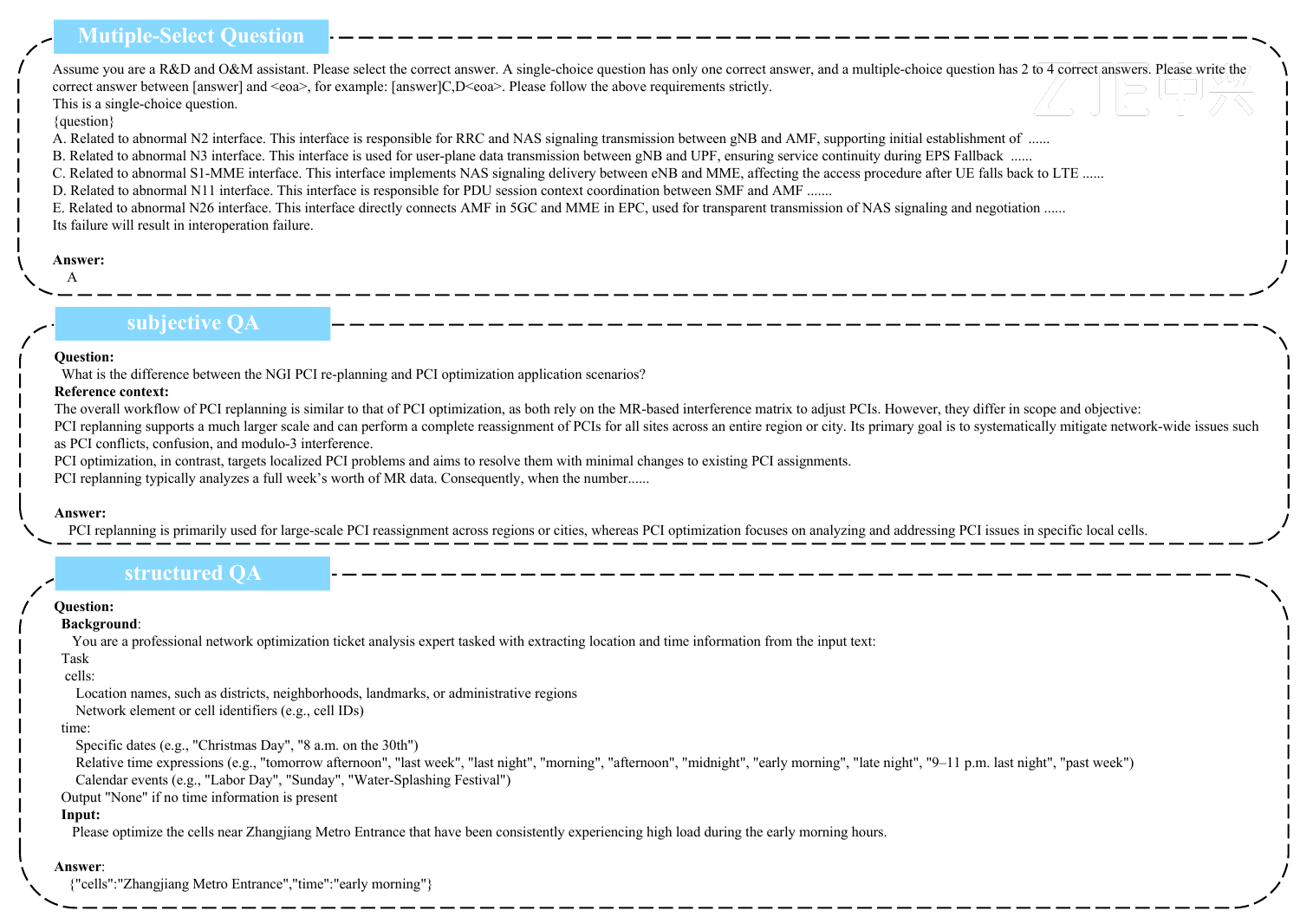}
\caption{Example formats of three main question types in \textsc{TeleCom-Bench}.}
\label{fig:question_format}
\end{figure*}

TeleCom-Bench is a benchmark dataset for evaluating professional knowledge and task-processing capabilities in the telecommunication domain. As shown in Figure~\ref{fig:question_format}, it includes three core question types with the following format specifications:

\begin{itemize}
    \item \textbf{Multiple-Select Question}: Tests knowledge of network architecture, interface functions, and protocols, supporting single-choice (1 correct answer) and multiple-choice (2–4 correct answers) formats. Answers are listed alphabetically between `` and `<eoa>` tags.
    \item \textbf{Subjective QA}: Requires open-ended responses to complex problems (e.g., network optimization), with answers being concise and logically consistent with reference context.
    \item \textbf{Structured QA}: An information extraction task that extracts location names, network element identifiers, and time expressions from text, outputting results in JSON format.
\end{itemize}

\begin{figure*}[h]

\centering
\includegraphics[width=1.0\linewidth]{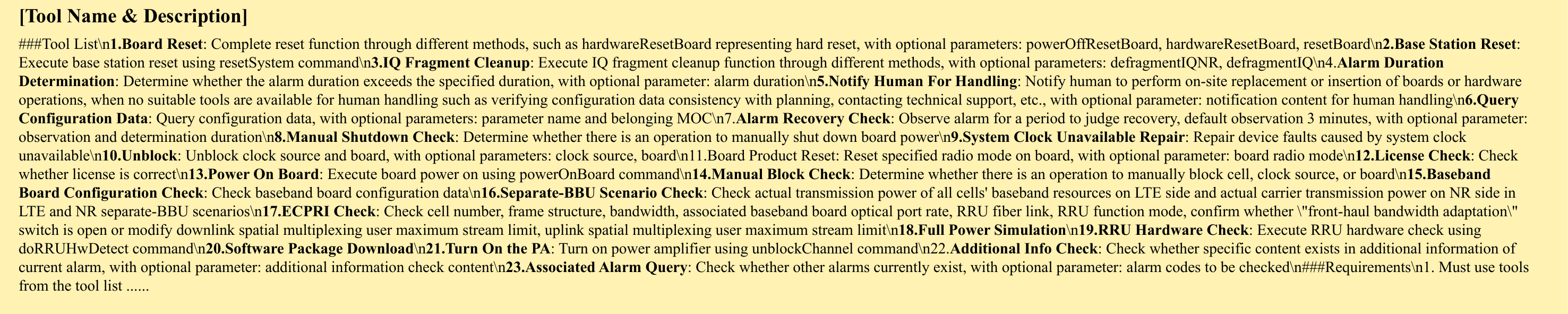}
\caption{
The 23-tool library provided as prompt context. Each tool includes a name and functional specification, enabling procedural reasoning. All models received this identical context.
}
\label{fig:detail_tools}
\end{figure*}

\subsection*{A.2 Detailed Experimental Breakdown Data}
As detailed in Table~\ref{tab:telecom_bench_detailed_performance}, we present a full breakdown of the experimental results reported in the main text.

\begin{table*}[htbp]
\centering
\caption{Detailed breakdown of model performance across task categories in \textsc{TeleCom-Bench}}
\label{tab:telecom_bench_detailed_performance}
\begin{adjustbox}{width=\linewidth,center}
\begin{tabular}{lllcccccccc}
\toprule
\multicolumn{3}{c}{Task Category} & Qwen3-32B & Qwen3-235B & DeepSeek-V3.2 & gemini2.5 & grok4.1 & GLM4.7 & doubao-pro 1.8 & kimi k2 \\
\midrule
\multirow{10}{*}{Knowledge Comprehension}
& \multirow{2}{*}{Basic Theory}
& Communication Exam & 86.77 & 90.77 & 89.41 & 86.43 & 86.69 & 80.31 & 91.40 & 52.58 \\
& & ieval-3GPP & 35.25 & 47.30 & 44.83 & 41.33 & 55.25 & 23.33 & 48.87 & 52.92 \\
\cmidrule{2-11}
& \multirow{3}{*}{Wired Network}
& Paragraph Summarization & 66.84 & 56.88 & 49.77 & 46.47 & 62.61& 56.95& 52.45 & 56.80 \\
& & Concept Explanation & 55.06 & 67.68 & 66.36 & 63.82 & 68.70 & 68.15& 65.60 & 63.80 \\
& & Simple QA & 56.33 & 58.09 & 55.90 & 53.74 & 56.49 & 57.62& 58.30 & 62.37 \\
\cmidrule{2-11}
& \multirow{2}{*}{Wireless Network}
& Network Optimization & 75.06 & 72.10 & 61.94 & 64.89 & 67.31 & 23.74& 61.99 & 64.81 \\
& & Fault Diagnosis & 71.01 & 69.49 & 29.83 & 62.38 & 37.84 & 21.51 & 28.24 & 75.53\\
\cmidrule{2-11}
& Core Network & CCN & 60.22 & 62.13 & 62.34 & 59.28 & 66.79 & 47.68 & 64.46 & 56.79 \\
\midrule
\multirow{18}{*}{Knowledge Application}
& \multirow{7}{*}{Intent Recognition}
& Fault Assistant\_Case Rec\_Intent & 100.00 & 100.00 & 100.00 & 100.00 & 95.69 & 99.52 & 100.00 & 100.00 \\
& & Fault Assistant\_Intent & 99.11 & 99.11 & 99.38 & 94.07 & 99.03 & 96.81 & 99.03 & 98.94 \\
& & Fault Assistant\_Scene & 99.05 & 97.78 & 97.78 & 97.78 & 97.47 & 94.30 & 97.78 & 97.78 \\
& & NetOpt Expert\_Agent Class & 94.69 & 100.00 & 100.00 & 100.00 & 100.00 & 100.00 & 100.00 & 100.00 \\
& & NetOpt Expert\_Coverage Agent & 85.00 & 82.00 & 79.00 & 77.00 & 87.00 & 82.00 & 80.00 & 75.00 \\
& & NetOpt Expert\_Capacity Agent & 82.35 & 92.00 & 86.00 & 90.00 & 89.00 & 84.00 & 86.00 & 92.00 \\
& & NetOpt Expert\_User Input\_Intent & 91.71 & 90.73 & 87.80 & 91.71 & 88.78 & 92.20 & 91.22 & 90.73 \\
\cmidrule{2-11}
& Entity Extraction & NetOpt Expert\_User Input\_Entity & 95.74 & 99.72 & 99.72 & 99.72 & 99.72 & 81.25 & 99.72 & 99.72 \\
\cmidrule{2-11}
& Tool Invocation & Smart Partner\_Workflow Gen & 84.71 & 45.54 & 94.06 & 90.80& 93.20& 56.50& 87.30& 84.50\\
\cmidrule{2-11}
& Event Verification &  4G Load Imbalance Check & 59.00 & 72.72 & 67.35 & 71.92 & 81.85 & 12.95 & 80.48 & 52.92 \\
\cmidrule{2-11}
& \multirow{2}{*}{Root Cause Diagnosis}
& Fault Agent & 60.00 & 99.52 & 82.86 & 59.8& 62.93 & 20.71 & 88.61 & 62.65 \\
& & Single-Domain Alarm Analysis & 61.83 & 43.46 & 43.13 & 38.76& 34.26 & 32.55&  34.04& 53.04\\
\cmidrule{2-11}
& \multirow{2}{*}{Fault Resolution}
& UME Fault Solution (EN) & 13.90 & 0.00 & 9.42 & 0.45 & 9.87 & 16.59 & 26.91 & 4.04 \\
& & UME Fault Solution (CN) & 16.14 & 0.00 & 1.79 & 44.39 & 19.28 & 2.69 & 34.53 & 12.56 \\
\bottomrule
\end{tabular}
\end{adjustbox}
\end{table*}

\subsection{A.3 Model Outputs and Tool Context: Procedural Reasoning Failures in Telecom Fault Resolution}

\begin{figure*}[h]
\centering
\includegraphics[width=1.0\linewidth]{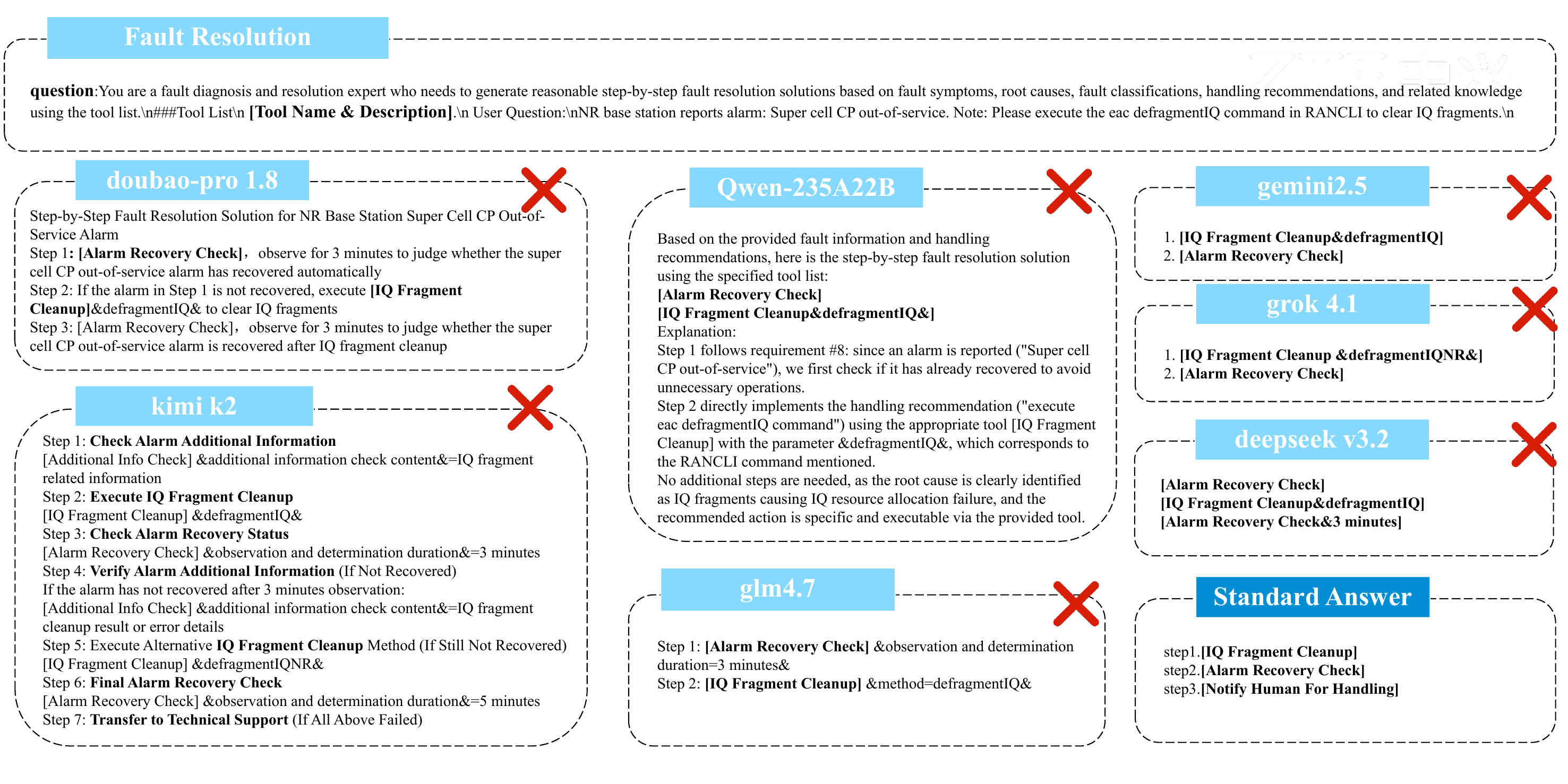}
\caption{
Model responses to a complex fault resolution task. Generalist models produce unstructured advice or hallucinated commands, failing to utilize the provided tool interface.
}
\label{fig:model_output}
\end{figure*}

Figures~\ref{fig:model_output} and~\ref{fig:detail_tools} illustrate a tool-augmented fault resolution scenario from \textsc{TeleCom-Bench}, focusing on the diagnosis of an ``NR Base Station Super Cell CP Out-of-Service'' alarm.
Figure~\ref{fig:detail_tools} presents the shared tool library provided to all models: 23 domain-specific functions (e.g., \texttt{[IQ Fragment Cleanup]}, \texttt{[RANCLI Executor]}, \texttt{[Alarm Correlation Analyzer]}), each with precise specifications of purpose, input/output formats, and usage constraints. This forms the uniform reasoning context for all models.
Figure~\ref{fig:model_output} shows the actual responses from seven generalist models (Qwen3-235A22B, DeepSeek-V3.2, Kimi-k2, GLM-4.7, Gemini-2.5, Grok-4.1, Doubao-Pro-1.8). Despite identical access to tool descriptions and task context, all generalist models fail to effectively utilize the tool interface: they either ignore the tool specifications and output unstructured natural language advice, or misuse tool semantics to generate non-executable procedures or hallucinated commands.
Together, these results demonstrate that the failure of generalist models in industrial telecom operations stems not from a lack of information, but from an inability to translate domain knowledge into structured, executable workflows—a critical requirement for fault resolution in telecom operations.


\end{document}